\documentclass[letterpaper, 10 pt, conferencek]{IEEEtran}  % Comment this line out if you need a4paper

\IEEEoverridecommandlockouts                              % This command is only needed if 
\usepackage{graphics} % for pdf, bitmapped graphics files
\usepackage{epsfig} % for postscript graphics files
\usepackage{amsmath} % assumes amsmath package installed
\usepackage{amssymb}  % assumes amsmath package installed
\usepackage{graphicx}
\usepackage[labelfont=bf,font=small]{caption}
\usepackage{multirow}
\usepackage{url}
\usepackage{rotating}
\usepackage{bm}
\usepackage{slashbox}
\usepackage{subfig}
\usepackage{color}
\usepackage{hyperref}
\title{\LARGE \bf
Deep Neural Ensemble for Retinal Vessel Segmentation in Fundus Images towards Achieving Label-free Angiography
}

\author{Avisek Lahiri, Abhijit Guha Roy, Debdoot Sheet, Prabir Kumar Biswas% <-this % stops a space
\thanks{Authors are with  Indian Institute of Technology Kharagpur, India.
        {\tt\small Email:  avisek@ece.iitkgp.ernet.in}}%%
}

\begin{document}

\maketitle
\thispagestyle{empty}
\pagestyle{empty}

%%%%%%%%%%%%%%%%%%%%%%%%%%%%%%%%%%%%%%%%%%%%%%%%%%%%%%%%%%%%%%%%%%%%%%%%%%%%%%%%
\begin{abstract}

Automated segmentation of retinal blood vessels in label-free fundus images entails a pivotal role in computed aided diagnosis of ophthalmic pathologies, viz., diabetic retinopathy, hypertensive disorders and cardiovascular diseases. The  challenge remains active in medical image analysis research due to varied distribution of blood vessels, which manifest variations in their dimensions of physical appearance against a noisy background.
 In this paper we formulate the segmentation challenge as a classification task. Specifically, we employ  unsupervised hierarchical feature learning using ensemble of two level of sparsely trained denoised stacked autoencoder.  First level training with bootstrap samples ensures decoupling and second level ensemble formed by different network architectures ensures architectural revision. We show that ensemble training of auto-encoders fosters diversity in learning dictionary of visual kernels for vessel segmentation. SoftMax classifier is used for fine tuning each member auto-encoder and multiple strategies are explored for 2-level fusion of ensemble members. On DRIVE dataset, we achieve maximum average accuracy of 95.33\%  with an impressively low standard deviation of 0.003 and Kappa agreement coefficient of 0.708 . Comparison with other major algorithms substantiates the high efficacy of our model. 

\end{abstract}

%%%%%%%%%%%%%%%%%%%%%%%%%%%%%%%%%%%%%%%%%%%%%%%%%%%%%%%%%%%%%%%%%%%%%%%%%%%%%%%%
\section{Introduction}

Segmentation and delineation of the articulated topology of retinal vessel network and structural attributes of vessels such as thickness, run length, tortuosity and branching patterns provide first level pathological cue for identifying various cardiovascular and opthalmogic diseases such as diabetes, hypertension, arteriosclerosis  \cite{survey1}. Automatic analysis of retinal vessel topography assists in developing robust screening systems for diabetic retinopathy \cite{survey2}, localizing of foveal avascular region \cite{survey4}, thinning of arteries \cite{survey5} and laser surgery \cite{survey1}. Vessel tortuosity embeds significant information about  hypertensive retinopathy \cite{survey6} while vessel diameter had been studied in connection with hypertension \cite{survey7}. Besides clinical areas, retinal topography has also been used for biometric applications \cite{survey11}.

\par \textbf{Related Works:} Unsupervised paradigms of automatic vessel detection primarily aims at designing matched filters by convolving a Gaussian kernel or its derivatives to emphasize vessel regions \cite{survey1}. Popular unsupervised multiscale approaches assign a `vesselness' metric based on eigen analysis of the Hessian \cite{survey12}. Unsupervised model based techniques include active contour models and geometric model guided level sets \cite{survey17} . Supervised models rely on feature extraction on manually annoted ground truth images followed by classification, usually using Artificial Neural Network (ANN),  Support Vector Machine (SVM) or their variants. Most common practise for feature extraction involves Gabor filter response at different strata of scale and degree \cite{survey20}. Other notable feature extraction methods include orientation analysis of gradient vector field \cite{survey6}, line operators \cite{survey21}, ridge profiles \cite{survey5} and modelling tissue-photon interactions \cite{debdootda}.
%===============  FIG    flowpart1   STARTS  =======================
\begin{figure}
\centering
\includegraphics[scale=0.42]{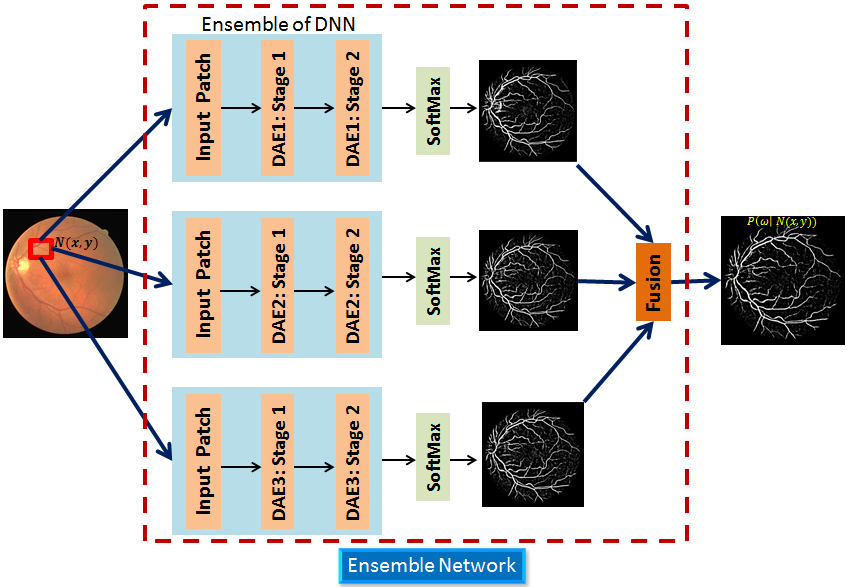}%prediction_figure.png}
\caption{Visualization of first level of ensemble learning. Each "E$_1$:Net" parallely train  $n$ denoising autoencoder based deep networks of depth $k$ on bootstrap training samples. Unsupervised feature extraction is followed by SoftMax classifier which produce probabilistic image maps which are finally conglomerated using different fusion strategies.}
\label{fig_flowpart1}
\end{figure}
%=============== FIG  flowpart1    ENDS  =======================
\par \textbf{Challenges and Mitigation Strategies:} The aforementioned traditional methods rely on hand crafted low level visual features and are inept at automatically learning generalized features from training data. Designing such carefully formulated task specific features requires appreciable domain expertise. Recent paradigms  of deep learning aim at unsupervised generalized feature extraction from the raw image data itself. Our work is primarily invigorated by the recent success of hierarchical feature extraction frameworks for retinal vessel segmentation \cite{anirban,abhijit} using deep neural networks.
%===============  FIG    flowpart2   STARTS  =======================
\begin{figure}
\centering
\includegraphics[scale=0.45]{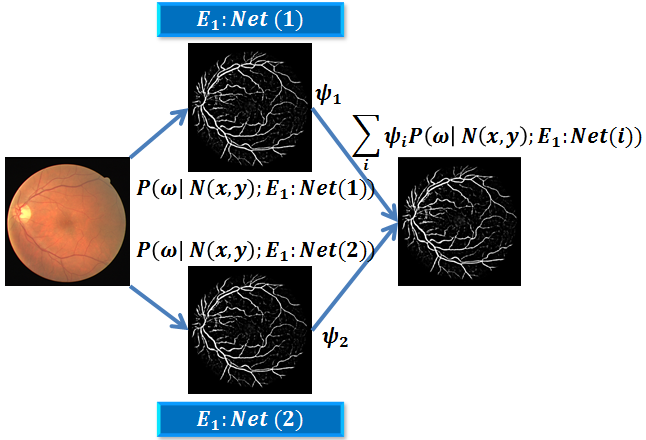}%prediction_figure.png}
\caption{Visualization of second level of ensemble learning. We train two parallel "E$_1$:Net"s of different architecture.}
\label{fig_flowpart2}
\end{figure}
%=============== FIG  flowpart2    ENDS  =======================
\par \textbf{Contributions:} 
 We present an architecture which supports two parallel levels of ensembling of stacked denoised autoencoder (SDAE) networks. As shown in Fig. \ref{fig_flowpart1}, first level of ensemble, also termed as "E$_1$:Net", is formed by training $n$ parallel SDAE networks (of same architecture) on booststrap samples while second level ensemble (Fig. \ref{fig_flowpart2}) is realized by parallel training of two "E$_1$:Net"s of different architecture. Initial architecture of any deep network is usually intuitive and needs to be fine tuned based on classification accuracy. We explore the possibility of leveraging performance of different architectured networks within the context of ensemble learning. We study various techniques for combining decisions of individual members of the ensemble and we show that using only simple SoftMax classifier we outperform the recent deep learning based method \cite{anirban} which uses random forest classifier after unsupervised feature extraction. For mitigating class imbalance between vessel and background, we present a set of smart sampling procedures for creating a database congenial for training SDAE.
\par The rest of the paper is organized as follows. The problem statement is formally defined in \S II and the
detailed solution is presented in \S III. In Section \S IV we validate and compare our results with state-of-the-art vessel segmentation algorithms. Finally, we conclude the paper in \S V with a summary of aptness of our ensemble learning of deep networks for automatically generating diversified dictionary kernels for medical image analysis.
%==========================================
\vspace{-2mm}\section{Problem Statement}
Given a color fundus image, $\mathcal{I} \in \mathcal{R}^{M\times N}$, our objective is to assign a probability to  each pixel (x,y) such that its neighbourhood, $N(x,y)$, centred at $(x,y)$ belongs to either of the classes, $\omega \in \{vessel, background\}$. Specifically, given a training set, $\{I_{train}\}$, we wish to formulate a function, $\mathcal H(\omega| N(x,y), \mathcal{I}; \{I_{train}\})$, whose response gives us $P(\omega | N(x,y))$. In this paper we learn the function $\mathcal H(.)$ by hierarchical feature extraction of $ \forall ~ N(x,y) \in \{I_{train}\}$ using our proposed fusion of SDAE ensemble networks.
%==========================================
\section{Proposed Approach}
\subsection{Image Preprocessing, sampling procedure and database formation}
Given a RGB fundus image, $\mathcal{I}$, we first extract the green channel image, $\mathcal{I}_g$, because it has been reported that the vascular structures manifest best contrast in green channel \cite{medi}. CLAHE is used for compensating irregular illuminations.
\par
\begin{figure}
\centering
\includegraphics[scale=0.4]{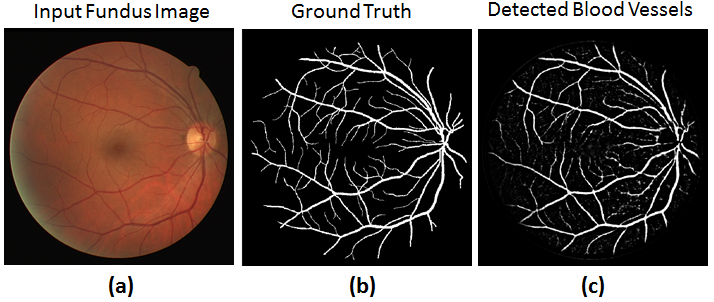}
\caption{Example of retinal vessel detection using our proposed algorithm on test image \# 19 of DRIVE \cite{drive} dataset.}
\label{fig_prediction_figure}
\end{figure}

 Fig. \ref{fig_prediction_figure} shows an exemplary manually annotated vessel network  used for extracting training patches. There exists significant class imbalance between background and vessel pixels. It is a challenging task to train classifiers in presence of highly skewed class distributions \cite{class_imbalance}. We propose a set of intuitive sampling strategies to mitigate this problem. Let  the ground truth binary image be $\mathcal{I}_{gr}$.  For sampling vessel patches, we first skeletonize the image $\mathcal{I}_{gr}$ to $\mathcal{I}_{gr}^s$. 
%In this regard we first compute the distance transform of $I_b$; image skeleton lies along the points of singularity, i.e., curvature discontinuities in the distance transform space.
 Image skeletonization helps in removing the boundary  foreground (vessel) pixels but retains the overall topological structure and thereby prevents redundant foreground sampling. Vessel training patches, $N(x,y)$, are uniformly sampled from $\mathcal{I}_g$ at those coordinates for which $\mathcal{I}_{gr}^s=1$. For sampling background pixels, we first morphologically dilate $\mathcal{I}_{gr}$ to $\mathcal{I}_{gr}^d$ by a square structural element of dimension $7 \times 7$. 
%Given an image, $A$ and structural element, $E$, image dilation, $A \oplus E$ is given by, 
%\begin{equation}
%A \oplus E= \bigcup\limits_{e \in E} (A+e)
%\end{equation}
Naively sampling at $\mathcal{I}_{gr}=0$ generates samples which are very near to original vessels and thus the neighbourhood encompass considerable region of vessel tissue region. Dilation operations thickens the vessels and thus sampling with $\mathcal{I}_{gr}^d$ ensures that the background training patches are well separated from original vessel regions. For each training image $\mathcal{I}$, if $V_I$ represents the set of vessel patches and $B_I$ represents set of background patches, we uniformly sample $|V_I|$ samples from $B_I$ to form $B^{'}_I$. As a final measure to enhance classification performance, an autoencoder is trained on a dataset with alternating patches from vessel and background class.
%========================
%===============  FIG    prediction_figure   STARTS  =======================
%=============== FIG  prediction_figure    ENDS  =======================
\vspace{-2mm}\subsection{Unsupervised layerwise prelearning using SDAE}
For automated feature discovery we have used stacked denoised autoencoders which are analogous in architecture to traditional multilayer perceptron (MLP) networks but in DAE, interconnection weights are learnt sequentially through unsupervised layerwise training (Refer to Sec. III-B \cite{SDAE} for more details). An autoencoder is basically a three layer fully connected network with one hidden layer which stores a compressed representation of input $\mathbf{x}$, and outputs, $\mathbf{\hat x}$, which is an approximation of $\mathbf{x}$. Let  $L$ denote number of layers, $s_l$ denote number of nodes in layer $l$, $W^{(l)}_{ji}$ denote weight between node $i$ of layer $l$ to node $j$ of layer $l+1$ and $h_{\mathbf{W}, \mathbf{b}}(.)$ is sigmoidal activation. According to \cite{SDAE}, loss function of sparse autoencoder is given by,\\\\
\vspace{-6mm}$$ L(\mathbf{W}, \mathbf{b})= \left [ \frac{1}{m}  \sum_{i=1}^m \left (   \frac{1}{2} ||  h_{\mathbf{W}, \mathbf{b}} (\mathbf{x_i})- \mathbf{x_i}    ||^2       \right )              \right ]  $$
\vspace{-2mm}\begin{equation}
+\lambda \sum_{l=1}^{L-1}  \sum_{i=1}^{s_l}  \sum_{j=1}^{s_{l+1}}(W^{(l)}_{ji})^2 ~+ \beta  \sum_{j=1}^{s_2} KL(\rho || \hat \rho)
\label{eq_optimization}
\end{equation}
The first term tries to minimize the discrepancy between input and predicted vector, second term is meant for L2 regularization to prevent over fitting  and the last term promotes sparsity within the network. $\beta$ is sparsity penalty, $m$ is cardinality of training sample space, $\rho$ is a sparseness parameter, $\bar \rho_k$  is th expected activation of node $k$ in hidden layer, i.e., $\bar \rho_k=\frac{1}{m}\sum_{i=1}^m z_k^{(i)}$, where, $z_k^{(i)}$ is the activation of node $k$ in hidden layer. DAE is a specialized version of autoencoder in which we incorporate  random additive noise at input side to transform $\mathbf{x}$ to $\mathbf{x}+\mathbf{r}$. Loss function in Eq. \ref{eq_optimization} is minimized by back propagation and the parameters $\mathbf{W}$ and $\mathbf{b}$ are updated using L-BFGS (Limited Memory Broyden-Fletcher-Goldfarb-Shanno). After this first step of pre-training, a stacked DAE is realized by treating the hidden layer node activations as input to  a second DAE and retraining the second DAE only. After second phase of pre-training we discard the last output layer and insert a simple SoftMax classification layer with only a single node which provides the probability, $P(\omega| N(x,y))$. Now, this SoftMax layer is used for fine tuning the entire stacked architecture similar to supervised MLP setting.\\
%========================
%===============  FIG    prediction_figure   STARTS  =======================
\begin{figure}
\centering
\includegraphics[scale=0.45]{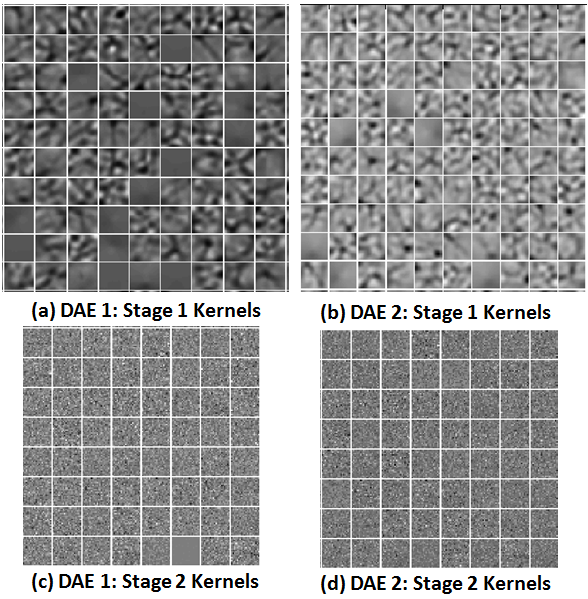}
\caption{Dictionary kernels learnt by $DAE_1$ and $DAE_2$ for `Ensemble Network 1'. We can clearly see that ensemble learning generates diversified kernels. Due to space constraints we refrain from showing the other kernels.}
\label{fig_kernels}
\end{figure}
%=============== FIG  prediction_figure    ENDS  =======================
\textbf{First level ensemble formation}
\label{sec_first_level}
Our first level of ensemble formation is inspired by success of bagging \cite{bagging} in ensemble learning. Given an original set, $Z$, of training examples, form $n$ sets, $Z_1$, $Z_2$,.. $Z_n$, each of cardinality $m$ by randomly sampling $m$ samples from $Z$ with replacement. We denote $DAE_ i$ as a SDAE of depth $k$ trained on $Z_i$ and collection of $n$ such SDAEs is termed as  "E$_1$:Net". In Fig. \ref{fig_flowpart1} and \ref{fig_kernels} and we have used $DAE_i$ and $DAE i$ interchangeably. As shown in Fig. \ref{fig_flowpart1}, during classification, each $DAE_ i$ produces a probabilistic output, $P(\omega|N(x,y); DAE_i)$. We used multiple strategies to fuse the probability maps from each $DAE_i$. 
\begin{equation}
P(\omega| N(x,y))_{min}= \underset{i} {\mathrm{argmin}}~P(\omega|N(x,y); DAE_i)
\label{eq_min}
\end{equation}
\begin{equation}
P(\omega| N(x,y))_{max}= \underset{i} {\mathrm{argmax}}~P(\omega|N(x,y); DAE_i)
\label{eq_max}
\end{equation}
\begin{equation}
P(\omega| N(x,y))_{avg}= \frac{1}{3}~\sum_{i=1}^3P(\omega|N(x,y); DAE_i)
\label{eq_avg}
\end{equation}
\vspace{-3mm}\begin{equation}
P(\omega| N(x,y))_{wavg}= \sum_{i=1}^3~\alpha_iP(\omega|N(x,y); DAE_i)
\label{eq_wavg}
\end{equation}
where $\alpha_k=\frac{r_k}{\sum_{j=1}^3r_j}$; $r_k$ is the accuracy rate of $DAE_k$ on cross validation set.\\
%=======================
\textbf{Second level ensemble formation}
 A SDAE (of depth 2) with SoftMax output is characterized by a $l-h_1-h_2-c$ network architecture, where $l=W\times W$ is input dimension, $c$ is number of classes and $h_1,h_2$ denote number of nodes in first and second hidden layer respectively. We induce further diversification in learning by parallel training of two " E$_1$:Net"s ( E$_1$:Net(1) and  E$_1$:Net(2)) of different architecture (refer to Fig. \ref{fig_flowpart2}), using different values of $h_1$ and $h_2$. Decisions from each "E$_1$:Net(i)" is merged by a convex weighted average; the weight being proportional to its accuracy on validation set.
%==============================================================================
\vspace{-3mm}\section{Results and Discussions}
We evaluate the performance of our algorithm on the popular DRIVE dataset \cite{drive} and compare our results with other state-of-the-art vessel segmentation methods. In our experiment we have used an architecture of `576-400-100-2' for " E$_1$:Net(1)" and `576-200-50-2' for " E$_1$:Net(2)". We have used $\lambda=0.001$, $\beta=3$, $W=24$; this setting yields the maximum accuracy averaged over test examples (Refer Table \ref{table_comparison}). Pre-training and finetuning of each SDAE is done for 700 epochs. 
\par In Fig. \ref{fig_kernels} we show some  exemplary visual dictionary kernels learnt by $DAE_1$ and $DAE_2$ of  E$_1$:Net. It is evident, specially for Stage 1, that training on bootstrap samples encourages the ensemble to learn diversified kernels. 
%Similar diversity of kernels is observed among the other DAEs. Each kernel is responsible for identifying a specific orientation of vessel. Learning diversified kernels thus enhances the representation prowess of our ensemble.
\par We use the two standard metrics, viz., maximum average accuracy and Cohen's Kappa agreement coefficient for comparing our results. In Table \ref{table_combinations} we first compare the performance of individual " E$_1$:Net" at first level of ensemble  using different fusion strategies as delineated in Sec. \ref{sec_first_level}.
%========================  table_combinations  STARTS=====================
\begin{table}
	\caption{Performances of individual "E$_1$:Net"  using different fusion strategies at first level of ensemble.}
	\centering
	\begin{tabular}{c l c c}\hline
		"E$_1$:Net" & Fusion  & \hspace{-6mm}Max. Avg. Accuracy & Kappa  \\\hline
		&Min (Eq. \ref{eq_min})& 0.932 & 0.687\\
                     1&Max (Eq. \ref{eq_max})& 0.928 & 0.679\\
                     &Average (Eq. \ref{eq_avg})& 0.908 & 0.654\\
                    &Weighted Average (Eq. \ref{eq_wavg})& 0.948 & 0.698\\\\
		&Min & 0.948 & 0.693\\
                     2&Max & 0.936 & 0.684\\
                     &Average & 0.910 & 0.667\\
                    &Weighted Average & 0.950 & 0.701\\\hline 
	\end{tabular}
	\label{table_combinations}
\end{table} 
%========================  table_combinations  ENDS=====================
%========================  table_comparison  STARTS=====================
\begin{table}
	\caption{Performance comparsion of competing algorithms.$(\mathbf{\sigma:})$ Standard deviation of max. avg. accuracy. Last four results are obtained from \cite{compare}.}
	\centering
	\begin{tabular}{ l c c}\hline
Method & Max. Avg. Accuracy ($\sigma$)& Kappa  Agreement\\\hline
                     \textbf{Proposed} & 0.953 (0.003)& 0.709\\
Second human observer & 0.947 (0.048)& 0.758\\
Maji et al. \cite{anirban} & 0.932~~~~~ (--)& 0.628\\
Roy et al. \cite{abhijit} & 0.912 (0.026)& 0.618\\
Sigurosson et al. \cite{prl}& 0.942 (0.010)& 0.708\\
Yin et al. \cite{medi} & 0.932 ~~~~~ (--)& -\\
Sheet et al. \cite{debdootda} & 0.976 ~~~~~ (--)& 0.821\\
Chaudhuri et al. & 0.877 (0.0232)& 0.33\\
Jiang et al. & 0.921 (0.0076)& 0.639\\
Martinez-Perez et al. & 0.918 (0.0240)& 0.638\\
Fraz et al. &0.948~~~~~(-)&-\\
Zana et al. \cite{last} & 0.937 (0.0077) & 0.697\\\hline
	\end{tabular}
	\label{table_comparison}
\end{table} 
%========================  table_comparison  ENDS=====================
We see that the weighted average method manifests best classification accuracy for both the first level ensemble networks. For fusing decision at second level of ensemble, we first generate the vessel probability maps by weighted average voting from both the level one "E$_1$:Net"s and average them to yield the final soft classification output (see Fig. \ref{fig_flowpart2}). Binarization of this posterior probability maps is achieved by thresholding the soft classification map at a level $L_t$ which maximizes the $F-Score$.
\par
\begin{figure}
\centering
\includegraphics[scale=0.4]{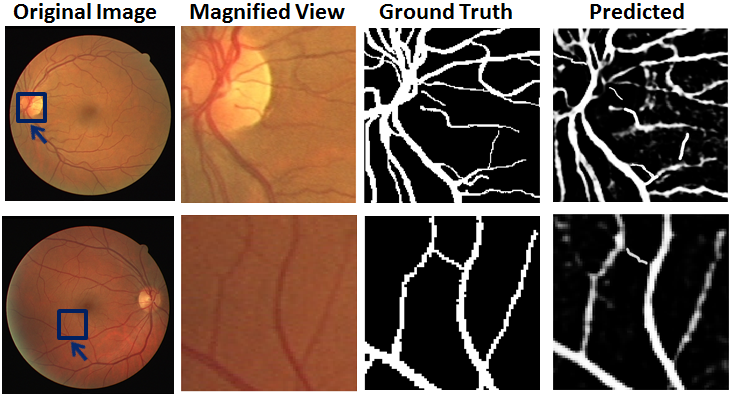}
\caption{Detection of course (top row) and fine (bottom row) vessels on two exemplary test images of  DRIVE dataset.}
\label{fig_thick_thin}
\end{figure}

 In Table \ref{table_comparison} we compare our results with state-of-the-art competing algorithms. Second column shows the maximum average accuracy along with the standard deviation. Comparison of Table \ref{table_combinations} and \ref{table_comparison} proves that second level ensemble learning further enhances the classification accuracy and thus justifies our approach.
 We achieve maximum average accuracy of 0.953 with standard deviation of only 0.003 and Kappa agreement coefficient of 0.709. It is encouraging to observe that our proposed unsupervised feature discovery based model superceeds the human observer in terms of accuracy and standard deviation.   Fig. \ref{fig_thick_thin} manifests the efficacy of our model in detecting both course and fine retinal vessels. Top row magnifies the optic nerve region where many vessels merge together and bundle up and segmentation becomes difficult but our algorithm performs appreciably in such region. Our proposed method also achieves high accuracy in segmenting sparsely distributed fine blood vessel (bottom row).
\vspace{-4mm}\section{Conclusion}
In this paper we presented a  deep neural ensemble network architecture for retinal vessel segmentation.  We have observed that miscellany of training space and architecture generated diversified dictionary of visual kernels for vessel detection. Each kernel is responsible for identifying a specific orientation of vessel. Learning diversified kernels thus enhances the representation prowess of our ensemble. Experimental validations suggest  that our unsupervised layerwise feature discovery based model is not only highly accurate but also reliable and consistent. Future improvements might focus on selecting a better threshold from the proabability maps. Another promising direction is to use multiview ensemble learning by extracting features from each stage of a stacked autoencoder.
%\vspace{-3mm}\section*{ACKNOWLEDGMENT}
%The work is financially supported by Google India PhD Fellowship awarded to the first author.
%===============  FIG    thick_thin   STARTS  =======================

%=============== FIG  thick_thin    ENDS  =======================
\bibliographystyle{IEEEtran}

\bibliography{IEEEabrv,ref3}

\begin{thebibliography}{10}
\providecommand{\url}[1]{#1}
\csname url@samestyle\endcsname
\providecommand{\newblock}{\relax}
\providecommand{\bibinfo}[2]{#2}
\providecommand{\BIBentrySTDinterwordspacing}{\spaceskip=0pt\relax}
\providecommand{\BIBentryALTinterwordstretchfactor}{4}
\providecommand{\BIBentryALTinterwordspacing}{\spaceskip=\fontdimen2\font plus
\BIBentryALTinterwordstretchfactor\fontdimen3\font minus
  \fontdimen4\font\relax}
\providecommand{\BIBforeignlanguage}[2]{{%
\expandafter\ifx\csname l@#1\endcsname\relax
\typeout{** WARNING: IEEEtran.bst: No hyphenation pattern has been}%
\typeout{** loaded for the language `#1'. Using the pattern for}%
\typeout{** the default language instead.}%
\else
\language=\csname l@#1\endcsname
\fi
#2}}
\providecommand{\BIBdecl}{\relax}
\BIBdecl

\bibitem{survey1}
J.~J. Kanski and B.~Bowling, \emph{Clinical ophthalmology: a systematic
  approach}.\hskip 1em plus 0.5em minus 0.4em\relax Elsevier Health Sciences,
  2011.

\bibitem{survey2}
T.~Teng, M.~Lefley, and D.~Claremont, ``Progress towards automated diabetic
  ocular screening: a review of image analysis and intelligent systems for
  diabetic retinopathy,'' \emph{Med. \& Biol. Eng. \& Comput.}, vol.~40, no.~1,
  pp. 2--13, 2002.

\bibitem{survey4}
A.~Haddouche, M.~Adel, M.~Rasigni, J.~Conrath, and S.~Bourennane, ``Detection
  of the foveal avascular zone on retinal angiograms using markov random
  fields,'' \emph{Digit. Signal Process.}, vol.~20, no.~1, pp. 149--154, 2010.

\bibitem{survey5}
E.~Grisan and A.~Ruggeri, ``A divide et impera strategy for automatic
  classification of retinal vessels into arteries and veins,'' in \emph{EMBC},
  vol.~1.\hskip 1em plus 0.5em minus 0.4em\relax IEEE, 2003, pp. 890--893.

\bibitem{survey6}
M.~Foracchia, E.~Grisan, and A.~Ruggeri, ``Extraction and quantitative
  description of vessel features in hypertensive retinopathy fundus images,''
  in \emph{Book Abstr. 2nd Intern. Workshop on Comput. Assist. Fundus Image
  Anal.}, vol.~6, 2001.

\bibitem{survey7}
J.~Lowell, A.~Hunter, D.~Steel, A.~Basu, R.~Ryder, and R.~L. Kennedy,
  ``Measurement of retinal vessel widths from fundus images based on 2-d
  modeling,'' \emph{IEEE Tran. Med. Imaging}, vol.~23, no.~10, pp. 1196--1204,
  2004.

\bibitem{survey11}
C.~Mari{\~n}o, M.~G. Penedo, M.~Penas, M.~J. Carreira, and F.~Gonzalez,
  ``Personal authentication using digital retinal images,'' \emph{Pattern Anal.
  \& Appl.}, vol.~9, no.~1, pp. 21--33, 2006.

\bibitem{survey12}
C.~K{\"o}se, C.~{\.I}ki \emph{et~al.}, ``A personal identification system using
  retinal vasculature in retinal fundus images,'' \emph{Expert Syst. with
  Appl.}, vol.~38, no.~11, pp. 13\,670--13\,681, 2011.

\bibitem{survey17}
R.~J. Winder, P.~J. Morrow, I.~N. McRitchie, J.~Bailie, and P.~M. Hart,
  ``Algorithms for digital image processing in diabetic retinopathy,''
  \emph{Computerized Med. Imaging \& Graphics}, vol.~33, no.~8, pp. 608--622,
  2009.

\bibitem{survey20}
M.~Niemeijer, J.~Staal, B.~Ginneken, M.~Loog, and M.~Abramoff, ``Drive: digital
  retinal images for vessel extraction,'' 2004.

\bibitem{survey21}
A.~Hoover, V.~Kouznetsova, and M.~Goldbaum, ``Locating blood vessels in retinal
  images by piecewise threshold probing of a matched filter response,''
  \emph{IEEE Tran. Med. Imaging}, vol.~19, no.~3, pp. 203--210, 2000.

\bibitem{debdootda}
D.~Sheet, S.~P.~K. Karri, S.~Conjeti, S.~Ghosh, J.~Chatterjee, and A.~K. Ray,
  ``Detection of retinal vessels in fundus images through transfer learning of
  tissue specific photon interaction statistical physics,'' in
  \emph{ISBI}.\hskip 1em plus 0.5em minus 0.4em\relax IEEE, 2013, pp.
  1452--1456.

\bibitem{anirban}
D.~Maji, A.~Santara, S.~Ghosh, D.~Sheet, and P.~Mitra, ``Deep neural network
  and random forest hybrid architecture for learning to detect retinal vessels
  in fundus images,'' in \emph{EMBC}.\hskip 1em plus 0.5em minus 0.4em\relax
  IEEE, 2015, pp. 3029--3032.

\bibitem{abhijit}
A.~Guha~Roy and D.~Sheet, ``Dasa: Domain adaptation for stacked autoencoders,''
  in \emph{ACPR}, 2015.

\bibitem{medi}
B.~Yin, H.~Li, B.~Sheng, X.~Hou, Y.~Chen, W.~Wu, P.~Li, R.~Shen, Y.~Bao, and
  W.~Jia, ``Vessel extraction from non-fluorescein fundus images using
  orientation-aware detector,'' \emph{Med. Image Anal.}, vol.~26, no.~1, pp.
  232--242, 2015.

\bibitem{class_imbalance}
M.~Galar, A.~Fernandez, E.~Barrenechea, H.~Bustince, and F.~Herrera, ``A review
  on ensembles for the class imbalance problem: bagging-, boosting-, and
  hybrid-based approaches,'' \emph{IEEE Tran. Syst. Man \& Cybernetics},
  vol.~42, no.~4, pp. 463--484, 2012.

\bibitem{drive}
J.~Staal, M.~D. Abr{\`a}moff, M.~Niemeijer, M.~A. Viergever, and
  B.~Van~Ginneken, ``Ridge-based vessel segmentation in color images of the
  retina,'' \emph{IEEE Tran. Med. Imaging}, vol.~23, no.~4, pp. 501--509, 2004.

\bibitem{SDAE}
P.~Vincent, H.~Larochelle, I.~Lajoie, Y.~Bengio, and P.-A. Manzagol, ``Stacked
  denoising autoencoders: Learning useful representations in a deep network
  with a local denoising criterion,'' \emph{JMLR}, vol.~11, pp. 3371--3408,
  2010.

\bibitem{bagging}
N.~A.~H. Mamitsuka, ``Query learning strategies using boosting and bagging,''
  in \emph{ICML}, vol.~1.\hskip 1em plus 0.5em minus 0.4em\relax Morgan
  Kaufmann Pub, 1998.

\bibitem{compare}
M.~Niemeijer, J.~Staal, B.~van Ginneken, M.~Loog, and M.~D. Abramoff,
  ``Comparative study of retinal vessel segmentation methods on a new publicly
  available database,'' in \emph{SPIE Med. Imaging}, 2004, pp. 648--656.

\bibitem{prl}
E.~M. Sigurosson, S.~Valero, J.~A. Benediktsson, J.~Chanussot, H.~Talbot, and
  E.~Stefansson, ``Automatic retinal vessel extraction based on directional
  mathematical morphology and fuzzy classification,'' \emph{Pattern Recognit.
  Lett.}, vol.~47, pp. 164--171, 2014.

\bibitem{last}
M.~M. Fraz, P.~Remagnino, A.~Hoppe, B.~Uyyanonvara, A.~R. Rudnicka, C.~G. Owen,
  and S.~A. Barman, ``An ensemble classification-based approach applied to
  retinal blood vessel segmentation,'' \emph{IEEE Tran. Biomed. Engg.},
  vol.~59, no.~9, pp. 2538--2548, 2012.

\end{thebibliography}

%%%%%%%%%%%%%%%%%%%%%%%%%%%%%%%%%%%%%%%%%%%%%%%%%%%%%%%%%%%%%%%%%%%%%%%%%%%%%%%%
\end{document}